\documentclass[runningheads]{llncs}

\usepackage{eccv}

\usepackage{eccvabbrv}

\usepackage{graphicx}
\usepackage{booktabs}

\usepackage[accsupp]{axessibility}  

\usepackage{hyperref}

\usepackage{orcidlink}

\begin{document}

\title{Deep Polarization Cues for Single-shot Shape and Subsurface Scattering Estimation} 

\titlerunning{Polarized Inverse Scattering}

\author{Chenhao Li\inst{1}\orcidlink{0009-0009-1594-5190} \and
Trung Thanh Ngo\inst{2}\orcidlink{0000-0002-7749-3726} \and
Hajime Nagahara\inst{1}\orcidlink{0000-0003-1579-8767}}

\authorrunning{Li et al.}

\institute{Osaka University, 2-8 Yamadaoka, Suita, Osaka, Japan \and
Hanoi University of Science and Technology, 1 Dai Co Viet, Hai Ba Trung, Hanoi, Vietnam
}

\maketitle

\begin{abstract}
  In this work, we propose a novel learning-based method to jointly estimate the shape and subsurface scattering (SSS) parameters of translucent objects by utilizing polarization cues. Although polarization cues have been used in various applications, such as shape from polarization (SfP), BRDF estimation, and reflection removal, their application in SSS estimation has not yet been explored. Our observations indicate that the SSS affects not only the light intensity but also the polarization signal. Hence, the polarization signal can provide additional cues for SSS estimation. We also introduce the first large-scale synthetic dataset of polarized translucent objects for training our model. Our method outperforms several baselines from the SfP and inverse rendering realms on both synthetic and real data, as demonstrated by qualitative and quantitative results.
  \keywords{Inverse Rendering \and Subsurface scattering \and Shape from Polarization}
\end{abstract}

\begin{figure}[t]
  \centering
  \includegraphics[width=\linewidth]{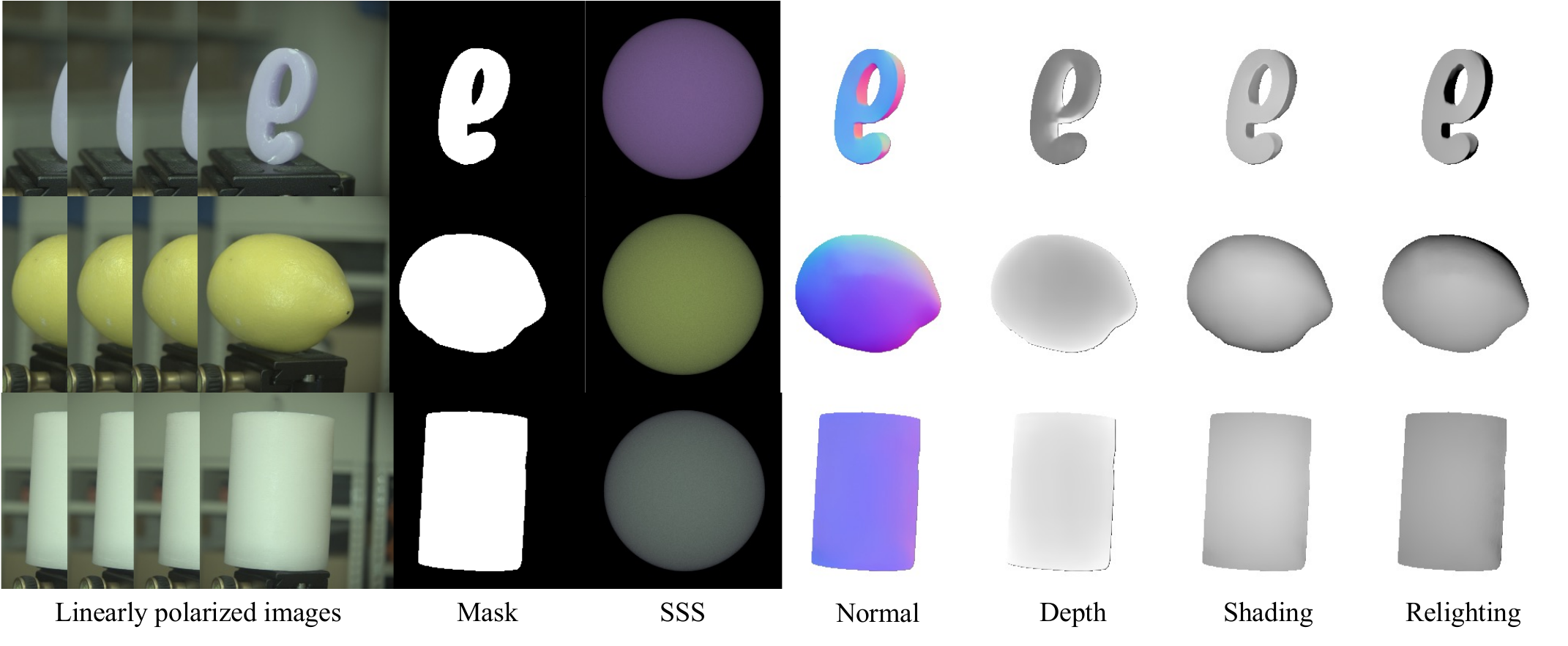}
  \caption{Visual results on real-world translucent objects. Our method takes four linearly polarized images and a binary mask as input. We use Mitsuba \cite{nimier2019mitsuba} to render a sphere to visualize estimated SSS parameters. For a better understanding of the quality of the estimated shape (Normal and Depth), we provide the shading and relighting results.}
  \label{fig:cover}
\end{figure}

\section{Introduction}
\label{sec:intro}

Translucent objects, such as human skin, milk, wax, and crystals, are omnipresent in our daily lives. A physical phenomenon that substantially influences the appearance of translucent objects is subsurface scattering. It occurs when photons penetrate the surface of an object and scatter inside it. After multiple scattering events, the photons are either absorbed by the object or exit from another surface point. Accurate estimation of subsurface scattering is crucial for various applications, including virtual reality, material science, and computer graphics.

However, estimating parameters for translucent objects is an ill-posed problem, even with the homogeneous assumption. In contrast to works \cite{boss2020two, li2018learning, deschaintre2021deep} that solely consider the interaction between light and the object on the surface, parameter estimation of translucent objects is more challenging since it requires accounting for both surface reflection and the multiple bounces and paths of SSS. Researchers often simplify scene representations to solve the parameter estimation problem of translucent objects. For instance, they may ignore surface reflection \cite{che2020towards, zheng2021neural} or use the simplified models such as Bidirectional Scattering Surface Reflectance Distribution Function (BSSRDF) to approximate optically thick materials in the form of complex surface reflection \cite{inoshita2014surface, dong2014scattering, yang2016inverse, deng2022reconstructing}, or dipole-diffusion approximation \cite{zhu2013estimating} that can only handle diffuse subsurface light transport.  To the best of our knowledge, using polarization cues for SSS parameter estimation in the presence of surface reflections has not been explored, and we make the first attempt in this work.

Light is an electromagnetic wave in which the wavelength determines its color, the amplitude determines its intensity, and the geometrical direction of oscillation leads to its polarization. While the human visual system is sensitive to color and intensity, for most of the natural scenes, we barely perceive polarization. Nevertheless, optical elements such as polarizers or wave plates can be used to measure the polarization of light. The measurement of polarization offers a plethora of computer vision applications. When light interacts with an object, the polarization changes by the surface orientation and material, which can be used as an additional cue to estimate the intrinsic factors of the object.

Polarization is a suitable cue for SSS estimation. From the theoretical view, we can roughly classify polarization into diffuse and specular polarization. Diffuse polarization arises from the summation of SSS and multiple bounces between microsurfaces, while specular polarization is produced from single-bounce surface reflection. The polarization signal of reflected light (specular polarization) is distinct from that of refracted light after multiple scattering (diffuse polarization). The reflected light is strongly polarized, particularly near Brewster's angle, whereas the light after multiple SSS is nearly unpolarized. SSS affects the ratio between diffuse and specular polarization, ultimately altering the captured polarization signal. This is the reason that polarization can be beneficial for SSS estimation. From the practical view, polarization cameras enable capturing four linearly polarized images in a single shot, making the capturing feasible even in non-laboratory settings.

The challenges of polarization for SSS estimation derive from various ambiguities. One of the ambiguities worth mentioning is called diffuse/specular polarization ambiguity. Translucent objects exhibit both types of polarization simultaneously, making it difficult to determine whether the captured polarization signal originates from diffuse polarization, specular polarization, or both. Most existing works \cite{riviere2017polarization, kadambi2015polarized, cui2017polarimetric} exclusively consider either specular or diffuse polarization. A few researchers consider both types of polarization simultaneously; however, these methods are either applicable only to opaque objects \cite{baek2018simultaneous} or use a simplified SSS model \cite{baek2022all} in the form of Gaussian distribution.

This paper proposes a learning-based approach for the simultaneous estimation of shape and SSS parameters using polarization cues. We employ several strategies to mitigate the aforementioned ambiguity issues. Our method uses four linearly polarized images (captured at polarizer angles of $0$, $\frac{\pi}{4}$, $\frac{\pi}{2}$, and $\frac{3\pi}{4}$) as input for each translucent object. We propose novel $max$ \& $ min$ polarization representations and also use them as additional inputs for reducing the ambiguity of surface reflection and SSS since the representations are related to the ratio of specular and diffuse polarization. Inspired by the previous work \cite{boss2020two} that step-by-step estimates shape and BRDF parameters to reduce ambiguities, we also train a stage-wise deep neural network that first estimates geometry and illumination, and then uses them as a guide for SSS estimation. We propose a novel reconstruction network to disambiguate SSS estimation. Using a differentiable renderer to reconstruct the scene is commonly used in previous inverse rendering works to improve the accuracy of estimation. However, there is no general-purpose differentiable renderer for the polarized inverse scattering, we bypass this issue by using a deep neural network to reconstruct the scene and compute a reconstruction network. To train our network, we constructed a large-scale synthetic dataset comprising 117K scenes. In each scene, we rendered a human-created 3D model with a specular BSDF and homogeneous SSS under environmental illumination using a polarized renderer. Our contributions can be summarized as follows:
\begin{itemize}
\item We introduce polarization cues to the deep learning domain for the joint estimation of shape and SSS parameters.
\item We propose $max$ \& $min$ polarization representations using an additional input as explicit cues for specular reflection and SSS separation.
\item We propose a reconstruction network for supervising SSS parameters from images.
\item We built a large-scale synthetic dataset of polarized translucent objects.
\end{itemize}

\section{Related Work}

\subsection{Inverse rendering}
Inverse rendering is a longstanding problem in computer vision that seeks to estimate material, shape, and illumination from single or multiple views. In this section, we will focus exclusively on single-view methods. We roughly divide these methods into three categories: pure surface reflection models, pure SSS models, and both.

So far, \textbf{pure surface reflection} models have dominated inverse rendering. These approaches assume that light interacts with objects only at the surface. Numerous studies have been published for estimating shape \cite{ranftl2020towards, fu2018deep, li2018deep}, BRDF \cite{aittala2016reflectance, aittala2015two, li2018materials, li2017modeling, deschaintre2018single, meka2018lime}, and illumination \cite{gardner2017learning, georgoulis2017around}, either separately or jointly \cite{sang2020single, wu2021rendering}. Li \etal \cite{li2018learning} achieved a milestone in this field by proposing a method that simultaneously estimates shape, illumination, and BRDF using deep neural networks. Later, some studies have shown that the performance could be further improved by using a flash and no-flash image pair  \cite{boss2020two}, a recursive network \cite{lichy2021shape}, polarization cues \cite{deschaintre2021deep}, or an RGB-D camera \cite{ku2022differentiable}. Meanwhile, some researchers have demonstrated that Li \etal  \cite{li2018learning} could be extended to scene-level inverse rendering by introducing a Residual Appearance \cite{sengupta2019neural} or spatially-varying illumination \cite{li2020inverse, Wang_2021_ICCV, zhu2022irisformer}. 

Another group of researchers has focused on \textbf{pure SSS} models. Such scene representations are often used to model participating media without clear boundaries, such as smoke, dust, or fog. Traditional methods \cite{levis2015airborne, gkioulekas2016evaluation, khungurn2015matching, zhao2016downsampling, gkioulekas2013inverse} were mainly based on analysis by synthesis. However, they suffered from many issues like long optimization time and local minimum. A recent breakthrough has been made by Che \etal \cite{che2020towards}. They combined a neural network and a differentiable renderer \cite{nimier2019mitsuba} and trained them end-to-end to estimate SSS parameters. 

Only a few works have considered \textbf{both surface reflection and SSS}, and it is still an open research topic. Our method belongs to this group. Baek \etal \cite{baek2022all} used a sophisticated temporal-polarimetric capturing system to simultaneously optimize the surface and subsurface parameters. However, the rendering model they used for SSS is a simple depolarization model in the form of Gaussian distribution. Additionally, they suffered from a long optimization time. Compared to their work, we use polarization to estimate SSS parameters. Besides, our method realizes the parameter estimation at an acceptable cost. The inference process is just a single forward propagation. At the same time, our capturing requirement is more flexible, allowing users to capture images without a laboratory environment. Li \etal \cite{li2023inverse} introduced a novel method that uses flash and non-flash image pairs to disambiguate the inverse scattering. Compared to their method, we allow the data capture within just a single shot and do not require additional lighting information.

\subsection{Polarization}
As noted, polarization is an active research area in computer vision and graphics. Numerous works have been reported for various applications such as SfP \cite{fukao2021polarimetric, ichikawa2021shape, ba2020deep, lei2022shape, shao2023transparent, chen2022perspective,zhu2019depth, kadambi2015polarized, zou20203d}, BRDF estimation \cite{deschaintre2021deep, hwang2022sparse, baek2018simultaneous, kondo2020accurate, zhao2022polarimetric}, depth estimation \cite{tian2023dps}, image segmentation \cite{qiao2023multi, kalra2020deep, mei2022glass, liang2022multimodal}, reflection removal \cite{lei2020polarized, lyu2019reflection, li2020reflection}, white balance \cite{ono2022degree}, pose estimation \cite{ cui2019polarimetric,zou2022human, gao2022polarimetric}, compass \cite{sturzl2017lightweight}, and sensor design \cite{kurita2023simultaneous}. This section only discusses some main topics highly relevant to our work. 

\textbf{SfP} has been extensively studied and can be considered a sub-task of our work. The theoretical basis of SfP comes from the Fresnel equations. When a beam of light interacts with the object's surface, the polarization changes according to the surface orientation and material of the object. Therefore, the measured polarization signal can be used for shape estimation. The challenges of SfP arise from various ambiguities. For example, the current capturing system cannot disambiguate the polarization signal for the Angle of Polarization $\theta$ and $\theta+\pi$. Also, the ambiguity between specular and diffuse polarization is another critical issue. To reduce the difficulty of the problem, researchers usually use a simplified polarization model. For example, most existing works \cite{riviere2017polarization, kadambi2015polarized, cui2017polarimetric} exclusively consider specular and diffuse polarization. Another research direction addresses ambiguity by combining polarization with cues such as shading \cite{smith2018height, mahmoud2012direct, ngo2015shape}, depth \cite{kadambi2015polarized}, or time-of-light \cite{baek2022all}. Recent breakthroughs have been dominated by deep learning, using convolutional networks to compensate for ambiguity. Ba \etal \cite{ba2020deep} proposed the first learning-based method for normal estimation. Lei \etal \cite{lei2022shape} extended their work to the scene-level normal estimation using a viewing encoding to handle the non-orthographic projection problem.

A recent trend in polarization research is the \textbf{joint estimation of normal and reflectance}. Baek \etal \cite{baek2018simultaneous, baek2020image} proposed a model that captures the polarimetric BRDF and surface normal simultaneously. The proposed method was later improved by utilizing the time-of-flight cues \cite{baek2022all}. Deschaintre \etal \cite{deschaintre2021deep} were the first to introduce the joint estimation of BRDF and surface normal to the realm of deep learning. Dave \etal \cite{dave2022pandora} successfully addressed the joint normal and BRDF estimation problem by using an implicit radiance field of polarization. Li \etal \cite{li2023neisf} solved the multiple bounced polarimetric light problem using a hybrid light path representation.

Finally, we discuss some polarization works on \textbf{challenging materials} such as transparent objects. The difficulty of dealing with transparent objects is that they inevitably exhibit reflection and refraction at the same time, and it is not easy to separate the reflection/refraction portion. Our target objects (translucent objects) face similar challenges to these works (transparent objects). Polarization is a suitable cue for solving such problems because the polarization signal for reflected and refracted light is different. Many existing works used this property for applications such as reflection removal \cite{lei2020polarized}, glass segmentation \cite{kalra2020deep}, and normal estimation for transparent objects \cite{shao2023transparent}.

\begin{figure}[t]
  \centering
  \includegraphics[width=0.9\linewidth]{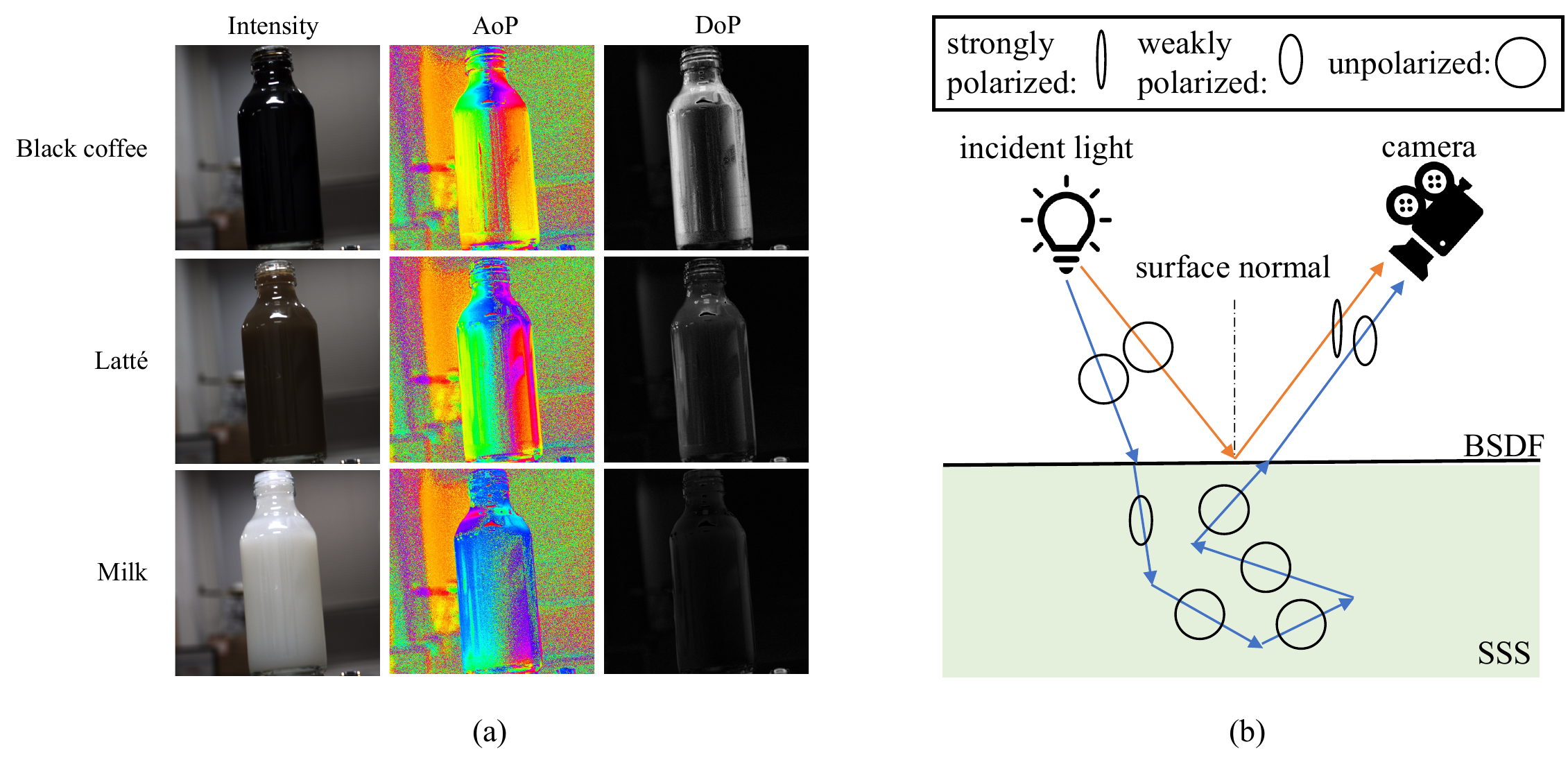}
  \caption{(a) SSS influences the polarization of objects. Images of a bottle (BSDF) were captured under the same illumination with three different liquids (SSS). For black coffee, most photons are absorbed upon entering the object, and the surface reflection dominates. For the milk, photons are back-scattered, and the SSS contributes significantly to the overall appearance. The latté falls somewhere in between these two extremes. (b) Our scene representation. We assume unpolarized light sources. The captured light intensity consists of two components. One comes from the single-bounce surface reflection, which exhibits \textbf{specular polarization} (orange path). Another one is the refracted light (blue path): An unpolarized light refracts into the object and becomes partially polarized light. Then, after undergoing the multi-bounce SSS, it becomes unpolarized light. Finally, the light leaves the object, undergoes Fresnel refraction again, and exhibits \textbf{diffuse polarization}.}
  \label{fig:concept}
\end{figure}

\section{Background}
\textbf{Light and optical element representation.}
 Following most SfP works \cite{ba2020deep, deschaintre2021deep, lei2022shape}, we only consider linear polarization. The linear polarization state of a beam of light is usually described by Stokes vectors $\mathbf{s}=[s_0, s_1, s_2]$. Where $s_0=I$ represents the unpolarized image intensity, $s_1$ represents the horizontal vs. vertical polarization, and $s_2$ represents the 45-degree polarization vs. the 135-degree polarization. Then the Degree of Polarization (DoP) $\rho$ and the Angle of Polarization(AoP) $\phi$ can be computed as follows:
\begin{equation}
    \rho = \frac{\sqrt{s_1^2+s_2^2}}{s_0},
    \label{equ:rho}
\end{equation}
\begin{equation}
    \phi = \frac{1}{2} \arctan \frac{s_2}{s_1}.
    \label{equ:phi}
\end{equation}
Mueller matrices $\mathbf{M} \in \mathbb{R} ^{3\times 3}$ are usually used to represent optical elements that change the polarization state. The final captured Stokes vector $\mathbf{s}_\text{out}$ is an integral over all possible light paths:
\begin{equation}
\mathbf{s}_\text{out} = \int_{\mathbb{P}}   f(\Bar{x}) \, d\Bar{x},
\end{equation}
where $\mathbb{P}$ is the space of all light paths, $\Bar{x}$ is a single path, and $f$ is the path contribution function.

\noindent \textbf{Our scene assumption.}
Following existing inverse rendering works \cite{li2018learning, boss2020two}, we assume the presence of two unpolarized light sources - natural lighting and a camera flashlight. The camera flashlight dominates the illumination. For the translucent object, we assume a smooth surface and a homogeneous SSS (SSS parameters are uniform within the object). A specular BSDF describes the object boundary, and similar to most SfP works \cite{ba2020deep}, we assume a constant Index of Refraction (IoR) of 1.5046 for the BSDF. To model the homogeneous SSS, we use the Radiative Transport Equation (RTE). The Henyey-Greenstein phase function \cite{HGphase1941} is used to represent the probability distribution of the outgoing direction when a photon interacts with particles inside the object. In theory, SSS depolarizes incident light based on the distance the light travels within the object \cite{germer2020evolution}. However, we assume that the light travels a sufficiently long distance and is already unpolarized before leaving the object's surface.

\noindent \textbf{Motivation of Polarization for SSS estimation.} We analyze two typical path contribution functions that majorly affect the captured Stokes Vectors $\mathbf{s}_\text{out}$. The first one is:
\begin{equation}
    f_r = \mathbf{M}_r \mathbf{s}_\text{in},
\end{equation}
where $\mathbf{s}_\text{in}$ is the Stokes vector of the light source, and $\mathbf{M}_r$ is the Mueller matrices of dielectric reflection. For convenience, we omit the rotation matrices here. This path describes the single bounce surface reflection (Orange path in Figure \ref{fig:concept} (b)) and exhibits \textbf{specular polarization}. The second one is:
\begin{equation}
    f_t = \mathbf{M}_t^o[\prod_i\sigma_t f_p(\theta_i, g)G(i, i+1)T(i, i+1)] \mathbf{M}_d \mathbf{M}_t^i \mathbf{s}_\text{in},
\end{equation}
where $\mathbf{M}_t^o$ and $\mathbf{M}_t^i$ are the Mueller matrices of dielectric transmission, $\mathbf{M}_d$ is a depolarizer, $\sigma_t$ is the extinction coefficient, $f_p$ is the phase function, $G$ is the geometry term, and $T$ is the transmission term. Refer to the blue path in Figure \ref{fig:concept} (b). $\mathbf{M}_t^o$ creates \textbf{diffuse polarization} when the light leaves the object's surface. SSS can affect the light intensity before $\mathbf{M}_t^o$, thus affecting $f_t$. The final polarization $\mathbf{s}_\text{out}$ is affected by the ratio between $f_t$ and $f_r$. Therefore, SSS can contribute to $\mathbf{s}_\text{out}$. In other words, $\mathbf{s}_\text{out}$ can be considered as a cue for SSS estimation. From Figure \ref{fig:concept} (a), we can observe that by switching the SSS (liquid in the bottle), the captured polarization signal changes significantly. Specifically, when the liquid is dark (the 1st row), most photons are absorbed during SSS. In this case, the reflected portion $f_r$ dominates, and the object exhibits specular polarization. When the liquid turns white (the 2nd and 3rd rows), the SSS portion $f_t$ increases and the object exhibits diffuse polarization. For most incident angles, the DoP of specular polarization is stronger than diffuse polarization. Therefore, we can observe that the DoP gets smaller as the liquid gets brighter. The AoP between the diffuse and specular polarization is $\frac{\pi}{2}$. Thus, the AoP also changes according to the SSS. This is why we bring polarization into the realm of SSS estimation.

\noindent \textbf{Polarization capturing.}
In our case, we use a polarization camera to capture four linearly polarized images ($I_0$, $I_{45}$, $I_{90}$, and $I_{135}$) in a single shot. The Stokes vectors can be calculated by the following equations:
\begin{equation}
    s_0 = \frac{1}{2}(I_0+I_{45}+I_{90}+I_{135}),
    \label{equ:unpol}
\end{equation}

\begin{equation}
    s_1 = I_{90} - I_{0},
\end{equation}

\begin{equation}
    s_2 = 2 * I_{45} - s_0.
\end{equation}
Then, we can obtain images with the maximum intensity $I_\text{max}$ and minimum intensity $I_\text{min}$:
\begin{equation}
    I_\text{max} = \frac{1}{2}(s_0 + \sqrt{s_1^2 + s_2^2}),
    \label{equ:max}
\end{equation}

\begin{equation}
    I_\text{min} = \frac{1}{2}(s_0 - \sqrt{s_1^2 + s_2^2}).
    \label{equ:min}
\end{equation}
We use these $max$ \& $min$ polarization representations because, for most incident angles, the specular polarization is stronger than the diffuse polarization. Therefore, when specular polarization dominates, the difference between $I_\text{max}$ and $I_\text{min}$ is significant, and we can observe a slight difference when diffuse polarization dominates. The network can learn the relationship between the specular and diffuse polarization and the $max$ \& $min$ images work for the separation cue between surface reflection and SSS components.

\begin{figure}[t]
  \centering
  \includegraphics[width=\linewidth]{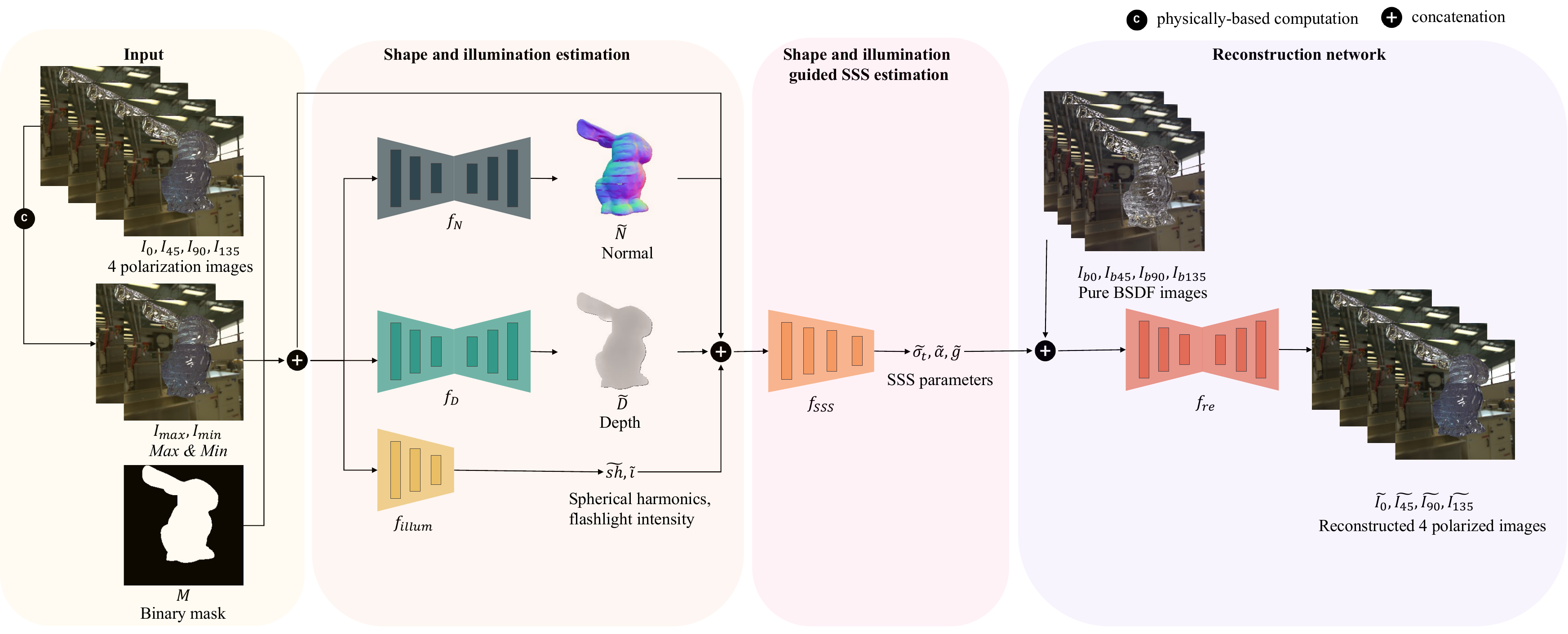}
   \caption{Overview of the proposed model. It takes four linearly polarized images ($I_0, I_{45}, I_{90}, I_{135}$), a $max$ \& $min$ physical prior ($I_\text{max}, I_\text{min}$), and a binary mask ($M$) as input. We use a stage-wise network architecture, and the model estimates shape and illumination first, then uses the estimated shape and illumination to guide the SSS parameter estimation. A novel reconstruction network whose inputs are four pure BSDF images ($I_{b0}, I_{b45}, I_{b90}, I_{b135}$) and the estimated SSS parameters is proposed to further reduce the ambiguity of SSS estimation. Note that the reconstruction network is only used during the training.}
  \label{fig:model_overview}
\end{figure}

\section{Our Model}
Given a translucent object with unknown illumination, shape, and material, our target is to estimate the shape and SSS parameters simultaneously. Similar to existing works \cite{boss2020two}, we use a depth map $D$ to represent coarse geometry and a normal map $N$ to provide local details. The illumination is also estimated as a side prediction to assist in the estimation. Following previous works \cite{li2018learning}, we use the first three-order spherical harmonics $sh$ to model the environment light and a flash intensity $i$ to determine the brightness of the camera flashlight. We model our SSS with three terms: a volumetric albedo $\alpha$, which determines the probability of whether photons are scattered or absorbed during a volume event; an extinction coefficient $\sigma_t$, which defines the optical density; and a Henyey-Greenstein phase function \cite{HGphase1941} parameter $g$ which controls whether the scattering is forward ($g>0$), backward ($g<0$) or isotropic ($g=0$). In summary, our \textbf{input} images are four linearly polarized images $I_0$, $I_{45}$, $I_{90}$, and $I_{135} \in \mathbb{R} ^{H \times W \times 3}$, two $max$ \& $min$ images $I_\text{max}$ and $I_\text{min}  \in \mathbb{R} ^{H \times W \times 3}$, and a binary mask $M  \in \mathbb{R} ^{H\times W}$ that locates the object. Our \textbf{output} parameters are:
\begin{itemize}
\item Shape: A normal map $N  \in \mathbb{R} ^{H\times W \times 3}$ and a depth map $D \in \mathbb{R} ^{H\times W}$ . 
\item Illumination: The spherical harmonics $sh  \in \mathbb{R} ^{3\times9}$ and flashlight intensity $i \in \mathbb{R} $.
\item SSS: The extinction coefficient $\sigma_t  \in \mathbb{R} ^{3}$, volumetric albedo $\alpha  \in \mathbb{R} ^{3}$, and Henyey-Greenstein phase function parameter $g  \in \mathbb{R}$. 
\end{itemize}
The existing works on BRDF estimation \cite{boss2020two} have used a cascaded network to address the shape/reflection ambiguity. Inspired by them, we also adopted this architecture. To be specific, we first estimate the shape and illumination, and then use the estimated parameters for SSS estimation. The model architecture is illustrated in Figure \ref{fig:model_overview}, and further details will be given in the following sections.

\subsection{Shape and illumination estimation}
For depth and normal estimation, we use a UNet-like network from a SOTA SfP work \cite{lei2022shape}. We concatenate four polarization images, $max$ \& $min$ images, and a binary mask in the channel dimension. The depth and normal are estimated via two separate networks $f_D$ and $f_N$:
\begin{equation}
    \Tilde{N} = f_N(P),
\end{equation}
\begin{equation}
    \Tilde{D} = f_D(P),
\end{equation}
where $P=(I_0, I_{45}, I_{90}, I_{135},I_\text{max},I_\text{min}, M)$ represents the input. The training losses of the estimated depth $\Tilde{D}$ and normal $\Tilde{N}$ are based on the $L_1$ distance between the estimated parameters and their GTs. 

We also estimate the illumination as a side prediction to assist with the SSS estimation. For the illumination network, we use a small CNN followed by a fully connected layer, with the same input as the depth and normal estimation. For the output, we use the first three-order spherical harmonics, which results in nine coefficients for each RGB channel. Combining these coefficients with the one-dimensional flash-light intensity, we have a total of 28 parameters for illumination:
\begin{equation}
    \Tilde{sh}, \Tilde{i} = f_\text{illum}(P).
\end{equation}
The illumination net $f_\text{illum}$ is trained using the $L_2$ loss between the estimated coefficients $\Tilde{sh}$ and flash light intensity $\Tilde{i}$ and their GTs.

\subsection{Shape and illumination guided SSS estimation}
\label{sec:subsurface scattering estimation}
The final appearance of an image is affected by the interplay of shape, illumination, and material. With the known shape and illumination, the ambiguity of SSS parameters estimation can be significantly reduced. Therefore, in addition to the original inputs, we also input the estimated shape and illumination information into the SSS net $f_\text{SSS}$:
\begin{equation}
    \Tilde{\sigma_t}, \Tilde{\alpha}, \Tilde{g} = 
    f_\text{SSS}(P, \Tilde{N}, \Tilde{D},\Tilde{sh}, \Tilde{i}),
\label{equ:SSS}
\end{equation}
where $\Tilde{\sigma_t}$, $\Tilde{\alpha}$, $\Tilde{g}$ are estimated extinction coefficients, volumetric albedo, and phase function parameters, respectively. We use a ResNet-like architecture for our SSS net. The SSS net is also trained by $L_2$ loss. Note that, similar to the idea of shape and illumination guided SSS estimation. Conversely, knowing the SSS can also disambiguate the shape or illumination estimation. Existing works \cite{nam2018practical, lensch2001image} used an iterative manner or cascaded networks \cite{boss2020two, li2018learning} to achieve this. For simplicity and computational efficiency, we only use one module, which is the shape and illumination-guided SSS estimation. However, the proposed model can be extended by cascading more X-guided X estimation modules.

\subsection{Reconstruction network}
In this section, we propose a novel reconstruction network to further disambiguate SSS estimation. As mentioned before, similar appearances can be achieved by different combinations of illumination, shape, and material, which makes inverse rendering extraordinarily challenging and ill-posed. To alleviate this problem, a typical approach is using a differentiable renderer \cite{che2020towards} to reconstruct the scene and compute the reconstruction network. However, there are several issues. First, to the best of our knowledge, there is no existing differentiable renderer that supports both surface reflection and SSS at the same time in polarized mode. While the latest work mitsuba3 \cite{Jakob2020DrJit} claims to support differentiable polarization, the differentiable polarization mode is incompatible with our scene representation (SSS+BSDF). Second, a well-known problem with differentiable renderers is the large memory usage and computational cost, and this problem is exacerbated in polarized mode. 

An advantage of using synthetic data is that we can create specific variations to the original dataset so that we can compute additional losses to train our model. Specifically, for each scene, we additionally render four pure BSDF polarized images ($I_{b0}$, $I_{b45}$, $I_{b90}$, and $I_{b135}$) by removing the SSS (see Figure \ref{fig:model_overview} for reference). Except for the SSS, the pure BSDF images and the original four polarized images have exactly the same illumination, shape, and BSDF. After Equation \ref{equ:SSS}, we input the estimated SSS parameters and pure BSDF images into a reconstruction network $f_\text{re}$ to reconstruct the original input images:
\begin{equation}
    \Tilde{I_{0}}, \Tilde{I_{45}},\Tilde{I_{90}},\Tilde{I_{135}} = f_\text{re}(I_{b0}, I_{b45}, I_{b90}, I_{b135}, \Tilde{\sigma_t}, \Tilde{\alpha}, \Tilde{g}).
\label{equ:reconstruction}    
\end{equation}
The reconstructed images are also penalized by the $L_1$ loss between their GT images $I_0$, $I_{45}$, $I_{90}$, and $I_{135}$. Our reconstruction network has the following merits: During reconstruction, the shape and illumination information is implicitly provided by the pure BSDF images, so there is less ambiguity of SSS. In addition, compared to the loss that is directly computed from a low-dimensional parameter space, a high-dimensional image space loss can provide more detail for gradient computing. Note that pure BSDF images are only required during training.

\section{Experiment}
\subsection{Training details}

The training was divided into two stages. We first trained the $f_N$, $f_D$, and $f_\text{illum}$ for 20 epochs using an Adam optimizer with a constant learning rate equal to 0.0002 for the first 10 epochs and a linearly decayed learning rate for the remaining 10 epochs. The batch size was set to 64. After that, we used the same approach to train the $f_\text{SSS}$ and $f_\text{re}$. Note that in the second stage, we did not freeze the weights of $f_N$, $f_D$, and $f_\text{illum}$. Instead, they continued to be trained. 

\subsection{Datasets}
In this section, we introduce the proposed synthetic dataset. A key issue in deep learning for inverse rendering is the lack of training datasets. Measuring the physical parameters of real-world objects, such as BRDF, SSS, and surface normals, can be time-consuming, especially with a large number of objects. With the rise of the computer graphics community, the current physically-based renderer can produce photorealistic images even for polarization. Therefore, many existing works \cite{deschaintre2021deep, li2018learning} use synthetic datasets for network training. 

In this work, we used Mitsuba 2 \cite{nimier2019mitsuba} for rendering the \textbf{synthetic dataset}. For 3D objects, we selected 5,847 human-created 3D objects from ShapeNet \cite{shapenet2015}. To create more variation of the surface normal of 3D objects, we collected 2,745 bump maps from several public sources. For the environment illumination, we used the Laval Indoor HDR dataset \cite{gardner2017learning}, which consists of 2,357 high-resolution indoor panoramas. Additionally, we also used a small sphere area light to mimic the camera flashlight. Each scene contains a randomly sampled 3D object, bump map, environment map, and SSS parameters. In total, we obtained 117K scenes, and each scene contains four linearly polarized images, four pure BSDF polarized images, and their GT parameters. All images' resolutions are 256$\times$256 with 4,096 samples per pixel. We used 100K for training and 17K for testing. To verify that the model trained on the synthetic dataset can be applied to the real-world translucent objects, we also constructed a \textbf{real-world dataset}. We calibrated our capturing system to mimic the synthetic data scene representation. For each object, we used a 3D scanner to obtain the shape and used Mitsuba to optimize the position to calculate the GT surface normals. We demonstrate the results on our real-world dataset in Figure \ref{fig:cover}. 
\begin{figure}[t]
  \centering
  \includegraphics[width=\linewidth]{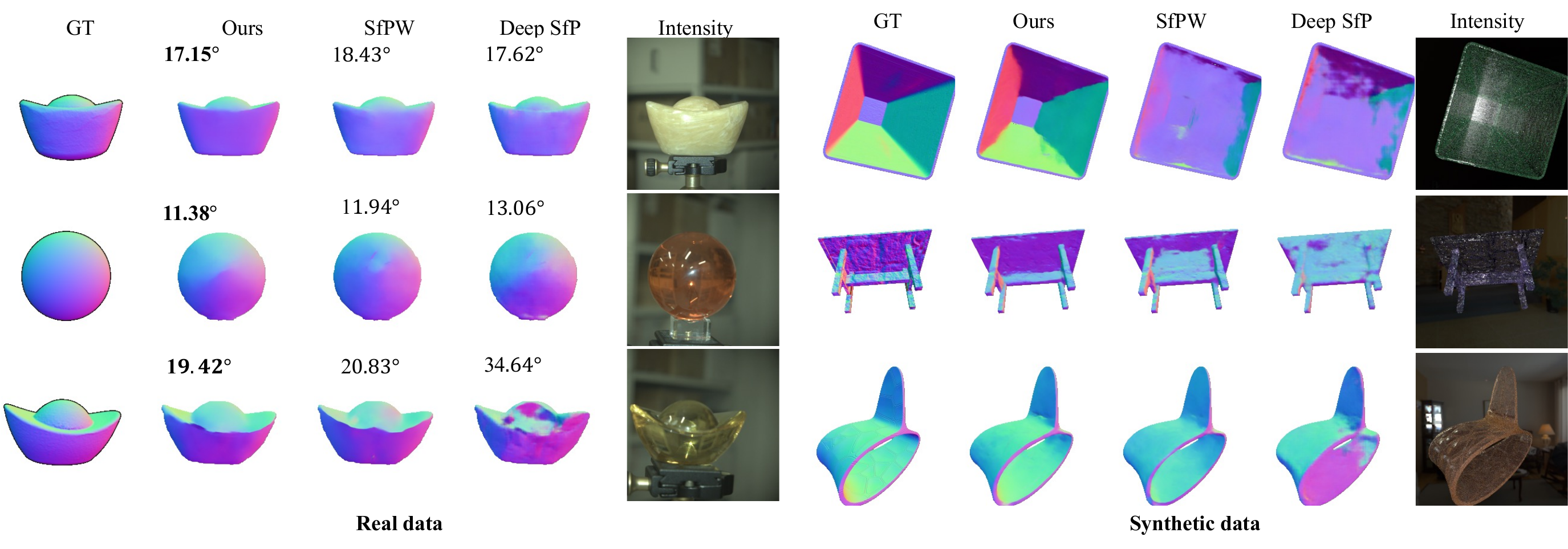}
  \caption{Visual comparison of estimated normal with Deep SfP \cite{ba2020deep} and SfPW \cite{lei2022shape}. Mean angular errors are provided on the top-left corners.}
  \label{fig:compare_sfp}
\end{figure}

\subsection{Comparison} 
We evaluate our approach by comparing it with the methods from the SfP and inverse rendering realm. During the following experiments, all methods are trained on our dataset with the same hyper-parameters.

\begin{table}[t]
  \centering
  \setlength{\tabcolsep}{5pt}
  \caption{MAE results on our 17K synthetic test dataset.}
  \begin{tabular}{*{6}c}
    \toprule
      &$N$ &$D$  &$\sigma_t$ &$\alpha$ &$g$ \\
    \midrule
    
    Che \etal \cite{che2020towards} &-	&-	&.1714	&.1026	&.2015\\
    Deep SfP \cite{ba2020deep} &.0831	&-	&-	&-	&-\\
    SfPW \cite{lei2022shape} &.0652	&-	&-	&-	&-\\
    \textit{w/o} Pol &.0750	&.0512	&.1398	&.0957	&.1717\\
    4 Pol   &.0657	&.0487	&.1369	&.0928	&.1602\\
    4 Pol + DoPAoP  &\textbf{.0641}	&.0486	&.1358	&.0969	&.1603\\
    4 Pol + MaxMin  &.0649	&.0483	&.1345	&.0931	&.1558\\
    \textbf{Ours (4 Pol + MaxMin + Re)} &.0651	&\textbf{.0473}	&\textbf{.1340}	&\textbf{.0918}	&\textbf{.1554}\\
    \bottomrule
  \end{tabular}
  \label{tab:mae}
\end{table}

\noindent \textbf{Comparison to SfP baselines.}
Since the recent progress of SfP was dominated by deep learning, we only compare our methods with learning-based SfP. Finally, We selected two methods which are Deep SfP \cite{ba2020deep} and SfPW \cite{lei2022shape}. \textbf{Deep SfP} is the first learning-based SfP method. They used four polarization images and pseudo-normal as input to train a UNet-like network. \textbf{SfPW} extended Deep SfP to the scene-level normal estimation by introducing a novel view encoding and an encoded AoP. We illustrate a visual comparison in Figure \ref{fig:compare_sfp} and report the Mean Absolute Error (MAE) in Table \ref{tab:mae}. It is observed that SfPW tends to estimate plane normals, while the results of Deep SfP are noisy. Our method can estimate high-quality normals for both synthetic and real data.

\begin{figure}[t]
  \centering
  \includegraphics[width=\linewidth]{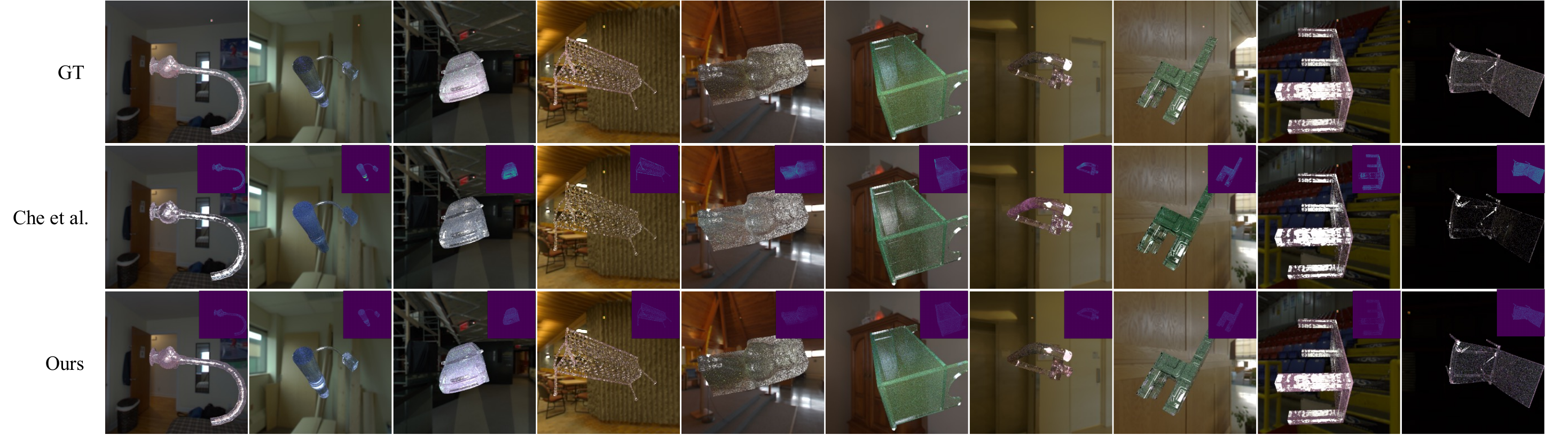}
  \caption{Visual comparison with an inverse scattering method. The 1st row is the GT intensity images. Images in the 2nd row are rendered by the GT shape, illumination, and estimated SSS by Che \etal \cite{che2020towards}. The 3rd row is the images rendered by SSS parameters estimated by our method. Error maps are provided in the upper right corner.}
  \label{fig:compare_sss}
\end{figure}

\noindent \textbf{Comparison to inverse rendering baselines.} It is not easy to find competitors for our method in the field of inverse rendering. Compared with \textbf{pure surface reflection models} is impossible. Because pure surface reflection models \cite{boss2020two, deschaintre2021deep, li2018learning} are usually modeled by BRDF, for our scene representation, we use BSDF to model the boundary and RTE to model the SSS. Thus, the parameters to be estimated are entirely different. Although Baek \etal \cite{baek2022all} also considered \textbf{both surface and SSS}, they only used a simplified SSS model for depolarization in the form of a Gaussian distribution. Although Li \etal \cite{li2023inverse} has a similar task as our method, comparing with them is unfair as they require additional lighting conditions and images as inputs. Therefore, it is not suitable to compare with their method. Finally, we selected a \textbf{pure SSS} method proposed by Che \etal \cite{che2020towards} to compare with our model. They trained an encoder-renderer model for SSS estimation, which can be considered a subtask of our work. Because it is not easy to compare the quality of estimated SSS without GT references of real-world objects, we only compare their method on the proposed synthetic dataset in Figure \ref{fig:compare_sss} and report MAE results in Table \ref{tab:mae}. Our method can estimate more accurate SSS parameters, and the images rendered by our estimated SSS parameters are close to the GT intensity images.

\subsection{Ablation study}
\begin{figure}[t]
  \centering
  \includegraphics[width=0.8\linewidth]{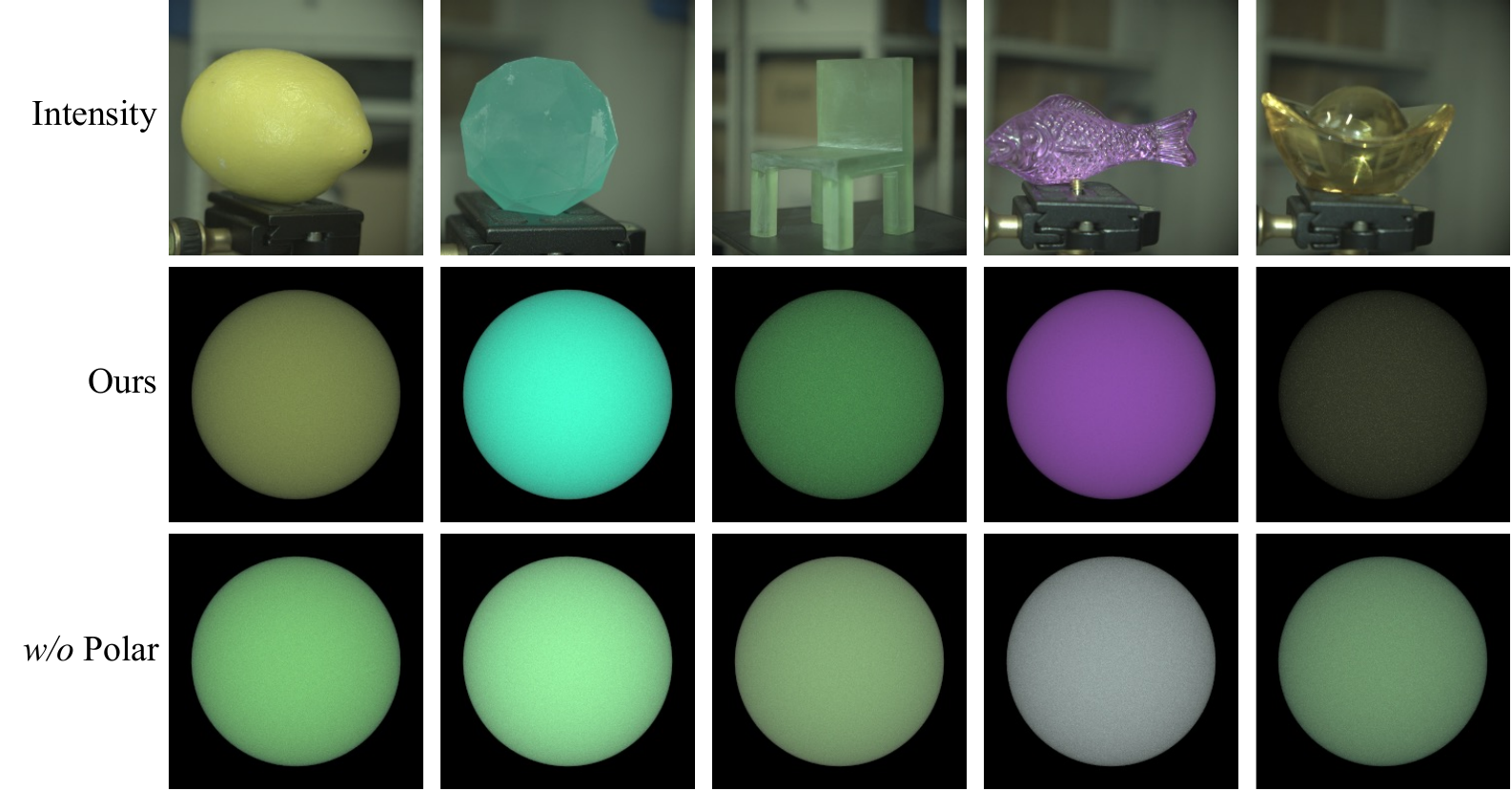}
  \caption{Visual comparison of SSS estimation results between our method and ``\textit{w/o} Polar".}
  \label{fig:compare_sss_real}
\end{figure}

In this section, we analyze the proposed method via several controlled experiments. We started from a naive baseline by removing the reconstruction network and using unpolarized images as input: $P=(I, M)$, where $I$ is the unpolarized light intensity. We call this experiment ``\textit{w/o} Polar". Next, instead of unpolarized images, we used four linearly polarized images: $P=(I_0, I_{45}, I_{90}, I_{135}, M)$, and denote this experiment as ``4 Pol". Then, we added AoP and DoP to our input: $P=(I_0, I_{45}, I_{90}, I_{135}, \rho, \phi, M)$, and used ``4 Pol + DoPAoP" to represent this experiment. After that, we replaced DoP and AoP images with the proposed $max$ \& $min$ images: $P=(I_0, I_{45}, I_{90}, I_{135}, I_\text{min}, I_\text{max}, M)$, and denoted this experiment as ``4 Pol + MaxMin". Finally, we added the reconstruction network (Equation \ref{equ:reconstruction}) back to obtain our full model. We compared these controlled experiments and reported MAE results in Table \ref{tab:mae}. From the table, we can observe that polarization cues significantly improved both shape and SSS estimation (See the comparison between ``\textit{w/o} Polar" and ``4 Pol"). Although ``4 Pol + DoPAoP" further improved the accuracy of shape estimation, the performance of scattering parameters was similar to ``4 Pol". The comparison between ``4Pol + MaxMin" and ``4 Pol + DoPAoP" demonstrated the efficiency of the proposed polarization representation in SSS estimation. Finally, the contribution of the proposed reconstruction network was validated by comparing our full model with the ``4 Pol + MaxMin" experiment. In addition, we provide a qualitative result of SSS estimation between our method and ``\textit{w/o} Polar" in Figure \ref{fig:compare_sss_real} to demonstrate the contribution of polarization cues.

\section{Conclusion and Limitation}
In this paper, we proposed the first method for joint estimation of shape and SSS parameters using polarization cues. We also constructed the first large-scale synthetic dataset of polarized translucent objects. The efficiency of the proposed $max$ \& $min$ polarization representation and reconstruction network were validated by several controlled experiments. In addition, we demonstrated that the proposed method outperforms the existing works from both SfP and inverse rendering realm, qualitatively and quantitatively.

However, we made several assumptions to reduce ambiguity. The first was a smooth surface to ensure the diffuse polarization only comes from SSS. However, if the object has a rough surface, the multiple reflections between microfacies also contribute to diffuse polarization. This can escalate ambiguity. Second, we assumed that photons travel far enough inside the object, which is inaccurate when the object has high transparency. Therefore, our method cannot be applied to nearly transparent objects. Solving these problems can be promising for future work. 

 \noindent \textbf{Acknowledgement} This paper is partially supported by JSPS KAKENHI 23H05490.

%
%
\bibliographystyle{splncs04}
\bibliography{main}
\end{document}